# Strategies to Improve Real-World Applicability of Laparoscopic Anatomy Segmentation Models

Fiona R. Kolbinger
Purdue University
`fkolbing@purdue.edu`

Jiangpeng He
Purdue University
`he416@purdue.edu`

Jinge Ma
Purdue University
`ma859@purdue.edu`

Fengqing Zhu
Purdue University
`zhu0@purdue.edu`

## Abstract

*Accurate identification and localization of anatomical structures of varying size and appearance in laparoscopic imaging are necessary to leverage the potential of computer vision techniques for surgical decision support. Segmentation performance of such models is traditionally reported using metrics of overlap such as IoU. However, imbalanced and unrealistic representation of classes in the training data and suboptimal selection of reported metrics have the potential to skew nominal segmentation performance and thereby ultimately limit clinical translation. In this work, we systematically analyze the impact of class characteristics (i.e., organ size differences), training and test data composition (i.e., representation of positive and negative examples), and modeling parameters (i.e., foreground-to-background class weight) on eight segmentation metrics: accuracy, precision, recall, IoU, F1 score (Dice Similarity Coefficient), specificity, Hausdorff Distance, and Average Symmetric Surface Distance. Our findings support two adjustments to account for data biases in surgical data science: First, training on datasets that are similar to the clinical real-world scenarios in terms of class distribution, and second, class weight adjustments to optimize segmentation model performance with regard to metrics of particular relevance in the respective clinical setting.*

## 1. Introduction

Semantic segmentation of anatomical structures and surgical instruments is an important component of holistic surgical scene understanding. In surgical data science [1,2], segmentation may be a component of complex downstream computer vision tasks including object-to-object interaction detection [3], action recognition [4], and surgical skill assessment [5,6]. Anatomy segmentation is an inherently challenging computer vision task: In contrast to structures with a static appearance like surgical instruments, anatomical structures may differ drastically in their visual appearance between patients (i.e., appearance of the liver surface in patients with and without hepatic cirrhosis) and within the same organ of one patient (i.e., ischemic and non-ischemic segments of the colon). As organs are also targets of a surgical intervention (i.e., organ resection, vessel ligation, anastomosis), their laparoscopic appearance may change substantially over the course of a surgical procedure. Technical and environmental conditions such as reflection artifacts and the presence of blood or smoke in an image further alter organs' visual appearance [7].

Available datasets to train and test segmentation models in surgical data science are often specifically curated for this technical purpose [8,9]. Correspondingly, they are composed to contain sufficient data of different classes rather than reflecting the true class distribution. This discrepancy between clinical reality and curated training and test data may result in models with limited clinical translatability despite good nominal performance indicated by classical technical metrics [10–12]. In this work, we present an empirical investigation of the effects of training data biases such as class imbalance and variations in modeling parameters in the specific use case of organ segmentation in laparoscopic image analysis. Based on a public benchmark dataset of laparoscopic images with corresponding organ segmentations [13], we illustrate the effects of the abovementioned variations on eight of the most commonly used segmentation metrics.

In summary, the contributions of our work are as follows:
— We illustrate the relations between training data composition as well as class (organ) size differences and segmentation performance in the context of organ segmentation in laparoscopic imaging. Our findings highlight the importance of critical evaluation of the data that models are trained and tested on.
— We demonstrate the effects of variations in class weights on organ segmentation performance in laparoscopy.
— By doing so, our work provides orientation as to the use of class weight adaptations to account for data biases in the context of organ segmentation in laparoscopy.

## 2. Related Works

### 2.1. Semantic Segmentation in Laparoscopic Imaging

Several teams have reported efforts in leveraging segmentation models for surgical scene understanding in laparoscopic imaging. A large proportion of works in the field have focused on instrument detection tasks, i.e., segmentation of surgical tools on bounding box level [15], detection and tracking of tool tips and other key points in two- and three-dimensional space [16], and instance segmentation [17–19]. A number of datasets are available for these tool-focused segmentation tasks [20–22].

In contrast, only few public datasets include detailed annotations of a range of anatomical structures, the most comprehensive of which is the Dresden Surgical Anatomy Dataset (DSAD) [13]. This dataset includes partially (binary) labeled data subsets for eleven organs and weak labels of organ presence for each image. The performance of basic segmentation models (DeepLabv3 and SegFormer) trained on DSAD was found to be comparable to human expert segmentation performance [14]. Jenke *et al.* have recently proposed integration of negative examples from data subsets with complementary organ annotations into training data for other classes as a data augmentation approach [23]. In addition to these newer approaches based on DSAD, multiple groups have previously investigated convolutional neural network (CNN)-based segmentation models for selected individual structures like liver, gallbladder, or uterus [8,24,25].

### 2.2. Strategies to Mitigate Data Bias and Class Imbalance in Semantic Segmentation

Data bias and class imbalance are common problems in computer vision that are most often caused by true differences in class frequencies, missing labels (i.e., when expert knowledge is required for annotations), or partially-labeled data. In deep learning, data bias may result in models overfitting to the majority class and performing poorly on underrepresented classes. In the context of this work, we primarily study the effects of foreground-background class imbalance (i.e., size differences between foreground and background). Several strategies have been proposed to mitigate the abovementioned challenges and their effects [26,27].

Resampling (i.e., oversampling of underrepresented classes and/or undersampling of overrepresented classes) to balance the dataset is a popular approach that can improve the performance of segmentation models. Examples for resampled training data can be identified in various ways including bootstrapping [28,29], Intersection-over-Union (IoU)-based sampling [30], or thresholding methods [31–33]. Modification of the loss function or introduction of class weights during training is another strategy to improve model performance on underrepresented classes. For example, focal loss [34] and class-balanced loss [35] have been proposed to mitigate the consequences of class imbalance in object detection and semantic segmentation tasks. Integration of synthetic data may help improve segmentation performance in class-imbalanced settings. In that regard, Zhao *et al.* have recently proposed integration of Masked Frequency Consistency to account for frequency variations between simulated data and real data influencing segmentation performance [36].

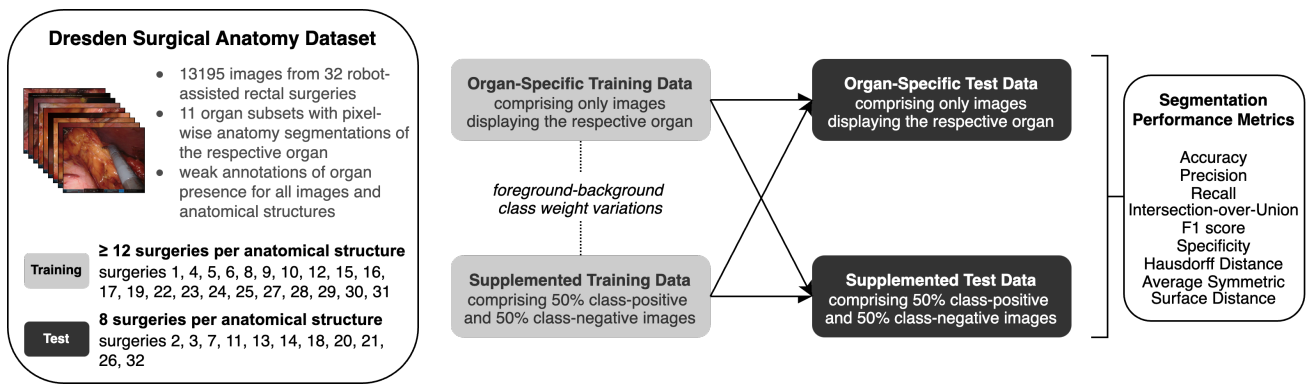

**Figure 1: Experimental setup.** Binary segmentation models were trained and tested on organ-specific or supplemented training data from the Dresden Surgical Anatomy Dataset providing comprehensive pixel-wise annotations of eleven abdominal organs in laparoscopic view.

## 2.3. Segmentation Performance Metrics

Accurate reporting of experimental results is the basis of reproducible and reliable science. In computational medical image analysis, selective reporting of segmentation metrics is a common shortcoming [37], which can result in limited reproducibility, clinical applicability and value of computational models [38–40]. Several works have described pitfalls pertaining to individual segmentation metrics [41–44], for example insensitivity of overlap-based metrics to structure boundaries or limited suitability of metrics penalizing both false positives and false negatives (i.e., IoU) for many use cases in medicine, where either false positives or false negatives should be minimized.

Based on a comprehensive summary of metric-related pitfalls in medical image analysis [41], Maier-Hein *et al.* have recently presented stakeholder consensus recommendations for medical image analysis metrics [45]. Generally, metrics should be appropriately selected to address the biomedical problem while considering metric-related pitfalls such as bias of the F1 score related to object size or shape. Recommended metrics for semantic segmentation include counting metrics such as F1 score (Dice Similarity Coefficient) and IoU as well as distance-based metrics such as Hausdorff distance (HD) and average symmetric surface distance (ASSD) [43,45]. Our work builds on these efforts and illustrates the consequences of selective reporting of metrics in the specific use case of organ segmentation in laparoscopic imaging.

## 3. Methodology

### 3.1. Dataset

For this work, we used the DSAD dataset [13], the most comprehensive publicly available laparoscopic image dataset with annotations of the presence and exact localization of abdominal organs. This dataset comprises 13195 distinct images, further subdivided into 11 organ subsets with at least 1,000 images from at least 20 patients each. Each organ subset contains binary annotations of the respective organ. In addition, the dataset contains binary weak labels for each image indicating the presence of all eleven organs in the respective image. Throughout this work, we followed the proposed split of surgeries into training, validation, and test data [14].

In terms of data for model training and testing, we compare two setups: (1) Organ-specific training and test data, which contain just class-positive images, as described in [14], and (2) supplemented training and test data, which comprise the organ-specific data and, in addition, an equal number of randomly sampled class-negative images that were sampled based on the weak labels provided in the DSAD (Figure 1). From a clinical perspective, the supplemented data better represent the clinical scenario of laparoscopic surgery, as it includes images with the respective target organ out of view. Therefore, we present the results of evaluation on the supplemented test dataset in the main manuscript, while the results from evaluation on the organ-specific test datasets are provided in the Supplementary Material.

### 3.2. Baseline Segmentation Method

SegFormer-based binary (structure-specific) segmentation models were implemented as described previously [14]. In brief, every model is a binary semantic segmentation model of a specific organ (such as the colon) and its corresponding background (non-colon). On an image y from the original dataset, the model's binary cross-entropy loss function $L_{orig}$, is given by:

$$L_{orig} = L_{pos} + L_{neg} \quad (1)$$

We split the loss function into two parts:

Positive Loss $L_{pos}$ (on foreground pixels, where $y_i = 1$):

$$L_{pos} = -\frac{1}{N}\sum_{i=1}^{N}[y_i \log(\hat{y}_i)] \quad (2)$$

Negative Loss $L_{neg}$ (on background pixels, where $y_i = 0$):

$$L_{neg} = -\frac{1}{N}\sum_{i=1}^{N}[(1-y_i)log(1-\hat{y}_i)] \quad (3)$$

Here, $N$ is the total number of pixels in the image from the original dataset. $y_i$ is the true label of the i-th pixel, where $y_i = 1$ indicates that the pixel belongs to the foreground, and $y_i = 0$ indicates that the pixel belongs to the background. $\hat{y}_i$ is the probability predicted by the model that the i-th pixel is part of the foreground.

Using this approach, $L_{orig}$ can guide the optimization of the model during the training phase. However, if the original dataset itself exhibits biases, such as training images always containing the target organ, or an imbalance in the number of pixels between the foreground and background, it is necessary to introduce modifications to the loss function.

### 3.3. Supplementary Negative Samples

All images in the original training dataset are selected from video frames of surgeries where the target organ appears within the camera's field of view, which does not reflect the variety encountered in actual surgeries.

Therefore, we introduce images of the same size that do not contain the organ from datasets of other abdominal organs as supplementary negative samples to address this bias. When the number of samples from the original dataset and the supplementary negative samples is at 1:1, the combined loss function, denoted as $L_{comb}$, can be formulated as:

$$L_{comb} = L_{orig} + L_{suppl} \quad (4)$$

$L_{suppl}$ can be written as:

$$L_{suppl} = -\frac{1}{N}\sum_{i=1}^{N}[\log(1 - \hat{z}_i)] \quad (5)$$

Here, $z$ represents a supplementary negative sample. $\hat{z}_i$ is the probability that the i-th pixel of $z$ is part of the foreground. The formats of (5) and (3) should be identical, but $z_i$ are 0 (meaning all pixels in $z$ belong to the background), the term $(1 - z_i)$ is omitted from the product.

### 3.4. Fore-Background Loss Weights

Severe fore-background imbalance in some organs' training datasets causes decaying performance on organs of smaller size or visible portion. In addition, supplementary negative training samples also result in a significantly higher number of negative pixels than positive pixels. These two points lead to positive loss $L_{pos}$ occupying only a small portion in $L_{comb}$, while the combined negative loss $L_{neg} + L_{comb}$ constitute a significant portion. This will result in the loss function not adequately reflecting the model's performance in the organ areas. To address this issue, we propose to re-weight the loss. Specifically, we introduce foreground loss weight $w_{pos}$ and background loss weight $w_{neg}$ and modify (4) to the following form that is used in our experiments:

$$L_{comb} = -\frac{w_{pos}}{N}\sum_{i=1}^{N}[y_i \log(\hat{y}_i)]$$
$$-\frac{w_{neg}}{N}\left(\sum_{i=1}^{N}[(1-y_i)\log(1-\hat{y}_i)] + \sum_{i=1}^{N}\log(1-\hat{z}_i)\right)$$
$$(6)$$

For extremely small organs such as pancreas and ureter, the value of $w_{pos}:w_{neg}$ should be greater than 1, thereby allowing the $L_{pos}$ to contribute a larger portion. However, it is worth noting that the optimal ratio of $w_{pos}:w_{neg}$ does not strictly equal the inverse ratio of the number of foreground to background pixels in the training datasets, because the fore-background loss weights also depend on other factors such as the difficulty of segmenting the organ or the quality of the training samples. Our experiments show that in specific surgical scenarios, if different metrics are focused on, the optimal choice of foreground-background loss weights also changes accordingly. In summary, the selection of foreground-background loss weights is empirical. We will analyze these scenarios in the next section in detail.

### 3.5. Segmentation Performance Metrics

We report six of the most commonly used and recommended [43,45] counting metrics for segmentation performance evaluation: accuracy, precision, recall, IoU [46], F1 score [47], and specificity. Counting metrics are calculated pixel-wise per image and are aggregated over all images in the test set. In addition, we report two distance-based metrics, HD and ASSD, calculated per image and aggregated over all images. Since these metrics require a ground truth annotation, we only report these metrics for the organ-specific test setups (Supplementary Material).

## 4. Experiments and Results

### 4.1. Organ Size Differences Relate to Organ Segmentation Performance

We first investigated the relationship between class (organ) segment size and segmentation performance metrics. Based on the average proportion of foreground pixels in the respective organ subset, organ sizes in the dataset range from 1.2% for the ureter to 26.2% for the abdominal wall. With increasing organ size, we observed a trend towards increasing F1 score, IoU, and recall, and slightly decreasing accuracy and specificity. Precision did not show a clear relation with organ size when trained on organ-specific training data. These tendencies pertained to models trained with class weights identical for foreground and background (Table 1) and organ size-adapted class weights (Table 2). We observed similar general relations between organ size and segmentation performance when models were trained on organ-specific training data and supplemented training data (Table 1, Table 2, Figure 2). When models were tested on organ-specific test data, we observed similar organ size-related trends for counting metrics. For the distance-based metrics (HD and ASSD), we did not find a clear relation with object size (Supplementary Table 1, Supplementary Table 2, Supplementary Figure 1).

### 4.2. Training Data Composition Impacts Organ Segmentation Performance

We next investigated the impact of training data composition on traditional segmentation performance metrics. When models were evaluated on test data containing both class-positive and class-negative images, training on data containing class-positive and class-negative examples resulted in a reduced frequency of false positives as compared to models trained on only class-positive examples. We consequently observed relative improvements in accuracy (particularly for large structures), specificity (particularly for large structures), and precision (particularly for small structures), with observed performance differences of up to 50% for precision of pancreas segmentation. Particularly for smaller structures, F1 score and recall decreased. For IoU, we observed no clear trend (Figure 2).

When models were tested on organ-specific test data not including negative examples, we observed overall similar trends related to training data composition. Addition of negative samples to the training dataset resulted in an increase of distance-based segmentation performance metrics (HD and ASSD) as compared to organ-specific training data (Supplementary Figure 1, Supplementary Table 1, Supplementary Table 2).

In most organs, the tendencies related to training data composition were considerably stronger when models were trained with 1:1 foreground-to-background class weights than with organ size-adapted class weights (Table 1, Table 2, Figure 2). Overall, these results illustrate that both training data composition and class weight variations considerably impact nominal segmentation performance, and that inclusion of class-negative training data improves accuracy, precision, and specificity of segmentation models.

**Table 1: Performance of binary organ segmentation models trained with class weights identical for foreground and background.** Models were trained on organ-specific (O) training data including only class-positive images or supplemented (S) training data including class-positive and class-negative examples. All models were tested on supplemented test data. Abbreviations: Intersection-over-Union (IoU), Organ-specific training dataset (O), Supplemented training dataset (S).

| Organ | Organ Size (% Foreground) | Accuracy | | Precision | | Recall | | IoU | | F1 Score | | Specificity | |
|---|---|---|---|---|---|---|---|---|---|---|---|---|---|
| | | O | S | O | S | O | S | O | S | O | S | O | S |
| Ureter | 1.2% | 99.51 | 99.43 | 58.26 | 42.71 | 24.92 | 19.94 | 21.14 | 15.73 | 34.91 | 27.19 | 99.9 | 99.86 |
| Intestinal Veins | 1.3% | 99.50 | 99.68 | 51.57 | 81.24 | 62.26 | 49.69 | 39.29 | 44.58 | 56.41 | 61.67 | 99.70 | 99.94 |
| Pancreas | 2.7% | 98.55 | 98.85 | 43.76 | 92.85 | 43.44 | 11.74 | 27.88 | 11.64 | 43.6 | 20.85 | 99.27 | 99.99 |
| Inferior Mesenteric Artery | 2.8% | 99.05 | 99.09 | 59.19 | 76.40 | 36.75 | 22.20 | 29.32 | 20.77 | 45.35 | 34.40 | 99.72 | 99.93 |
| Vesicular Glands | 2.9% | 98.37 | 98.56 | 43.88 | 74.06 | 30.33 | 6.36 | 21.85 | 6.22 | 35.87 | 11.71 | 99.41 | 99.97 |
| Spleen | 3.2% | 99.15 | 99.00 | 71.25 | 67.74 | 78.75 | 72.79 | 59.76 | 54.05 | 74.81 | 70.17 | 99.48 | 99.43 |
| Stomach | 5.0% | 94.55 | 97.28 | 27.21 | 48.21 | 63.61 | 41.36 | 23.55 | 28.64 | 38.12 | 44.52 | 95.38 | 98.79 |
| Colon | 11.8% | 93.71 | 94.66 | 46.64 | 52.56 | 72.68 | 63.02 | 39.69 | 40.17 | 56.82 | 57.32 | 94.98 | 96.57 |
| Small Intestine | 15.5% | 93.99 | 95.85 | 52.91 | 63.66 | 91.14 | 88.22 | 50.33 | 58.67 | 66.96 | 73.96 | 94.19 | 96.39 |
| Liver | 19.7% | 97.27 | 96.73 | 78.78 | 77.71 | 70.46 | 58.58 | 59.22 | 50.15 | 74.39 | 66.80 | 98.87 | 99.00 |
| Abdominal Wall | 26.2% | 94.69 | 95.58 | 77.79 | 84.65 | 81.10 | 79.44 | 65.85 | 69.44 | 79.41 | 81.97 | 96.65 | 97.92 |

**Table 2: Performance of binary organ segmentation models trained with class weights adapted to organ size.** Models were trained on organ-specific (O) training data including only class-positive images or supplemented (S) training data including class-positive and class-negative examples. All models were tested on supplemented test data. Abbreviations: Intersection-over-Union (IoU), Organ-specific training dataset (O), Supplemented training dataset (S).

| Organ | Organ Size (% Foreground) | Accuracy | | Precision | | Recall | | IoU | | F1 Score | | Specificity | |
|---|---|---|---|---|---|---|---|---|---|---|---|---|---|
| | | O | S | O | S | O | S | O | S | O | S | O | S |
| Ureter | 1.2% | 99.20 | 99.10 | 27.49 | 27.19 | 30.65 | 41.52 | 16.95 | 19.66 | 28.99 | 32.86 | 99.57 | 99.41 |
| Intestinal Veins | 1.3% | 98.92 | 99.52 | 29.80 | 52.79 | 79.29 | 79.52 | 27.65 | 46.47 | 43.32 | 63.45 | 99.03 | 99.63 |
| Pancreas | 2.7% | 98.62 | 99.08 | 47.26 | 76.73 | 61.38 | 41.27 | 36.43 | 36.68 | 53.40 | 53.67 | 99.10 | 99.84 |
| Inferior Mesenteric Artery | 2.8% | 98.89 | 98.90 | 48.26 | 49.00 | 45.77 | 44.42 | 30.70 | 30.38 | 46.98 | 46.60 | 99.47 | 99.50 |
| Vesicular Glands | 2.9% | 97.95 | 98.18 | 33.74 | 37.50 | 37.98 | 31.13 | 21.75 | 20.50 | 35.73 | 34.02 | 98.86 | 99.21 |
| Spleen | 3.2% | 98.62 | 98.95 | 54.55 | 63.79 | 84.77 | 79.93 | 49.68 | 54.98 | 66.38 | 70.95 | 98.84 | 99.26 |
| Stomach | 5.0% | 92.69 | 92.38 | 22.2 | 22.95 | 70.65 | 80.03 | 20.33 | 21.70 | 33.79 | 35.67 | 93.29 | 92.71 |
| Colon | 11.8% | 92.56 | 93.31 | 41.79 | 44.89 | 78.25 | 77.39 | 37.44 | 39.69 | 54.48 | 56.83 | 93.42 | 94.27 |
| Small Intestine | 15.5% | 95.42 | 94.34 | 60.88 | 54.45 | 88.14 | 93.48 | 56.27 | 52.46 | 72.02 | 68.82 | 95.94 | 94.40 |
| Liver | 19.7% | 96.91 | 96.65 | 70.30 | 75.08 | 77.91 | 60.46 | 58.61 | 50.35 | 73.91 | 66.98 | 98.04 | 98.80 |
| Abdominal Wall | 26.2% | 94.41 | 95.52 | 74.36 | 84.48 | 85.09 | 79.03 | 65.78 | 69.01 | 79.36 | 81.66 | 95.76 | 97.90 |

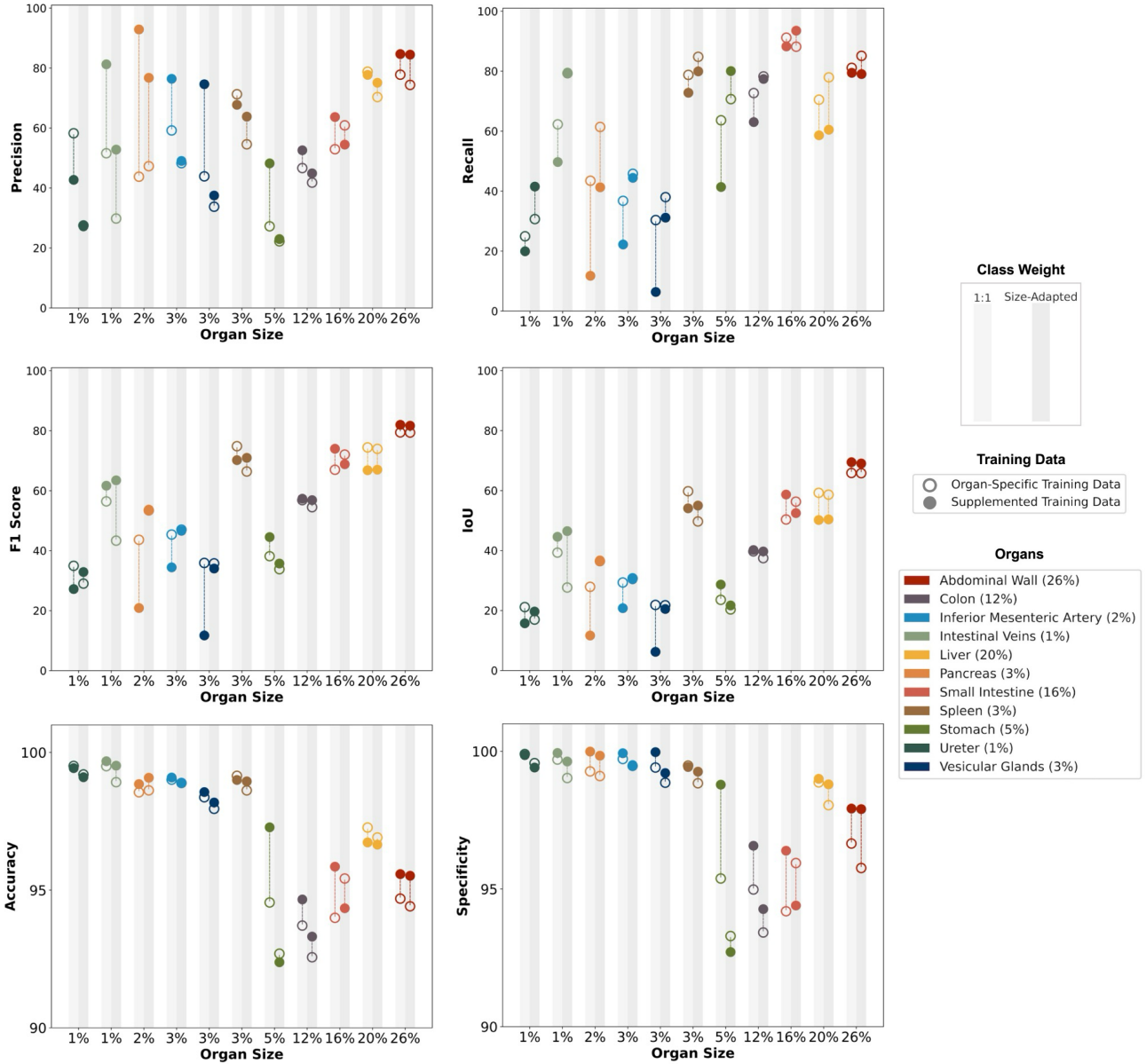

**Figure 2: Quantitative evaluation of segmentation model performance in relation to training data composition and foreground-background class weight.** Models were trained on either organ-specific training data including only class-positive images or supplemented training data including class-positive and class-negative examples. Foreground and background were either weighted identically (1:1) or adapted to organ size. Models were tested on supplemented test data. Abbreviation: Intersection-over-Union (IoU).

### 4.3. Adaptation of Class Weights to Optimize Segmentation Performance in Specific Use Cases

Building on our observation of the relevance of class weight changes, we systematically investigated the impact of varying the foreground-to-background class weight ratio between 0.7 and 15 on segmentation performance metrics. Variations in class weights impacted nominal segmentation performance for all organs across metrics. For individual metrics, we observed distinct trends that were partially related to organ size: Accuracy and specificity decreased with increasing class weights, particularly for large organs. Precision decreased for all organs with increasing class weights, independent of organ size. Recall increased with increasing class weights, particularly for smaller organs. In comparison to the other counting metrics, F1 score and IoU remained more stable when class weights varied (Figure 3).

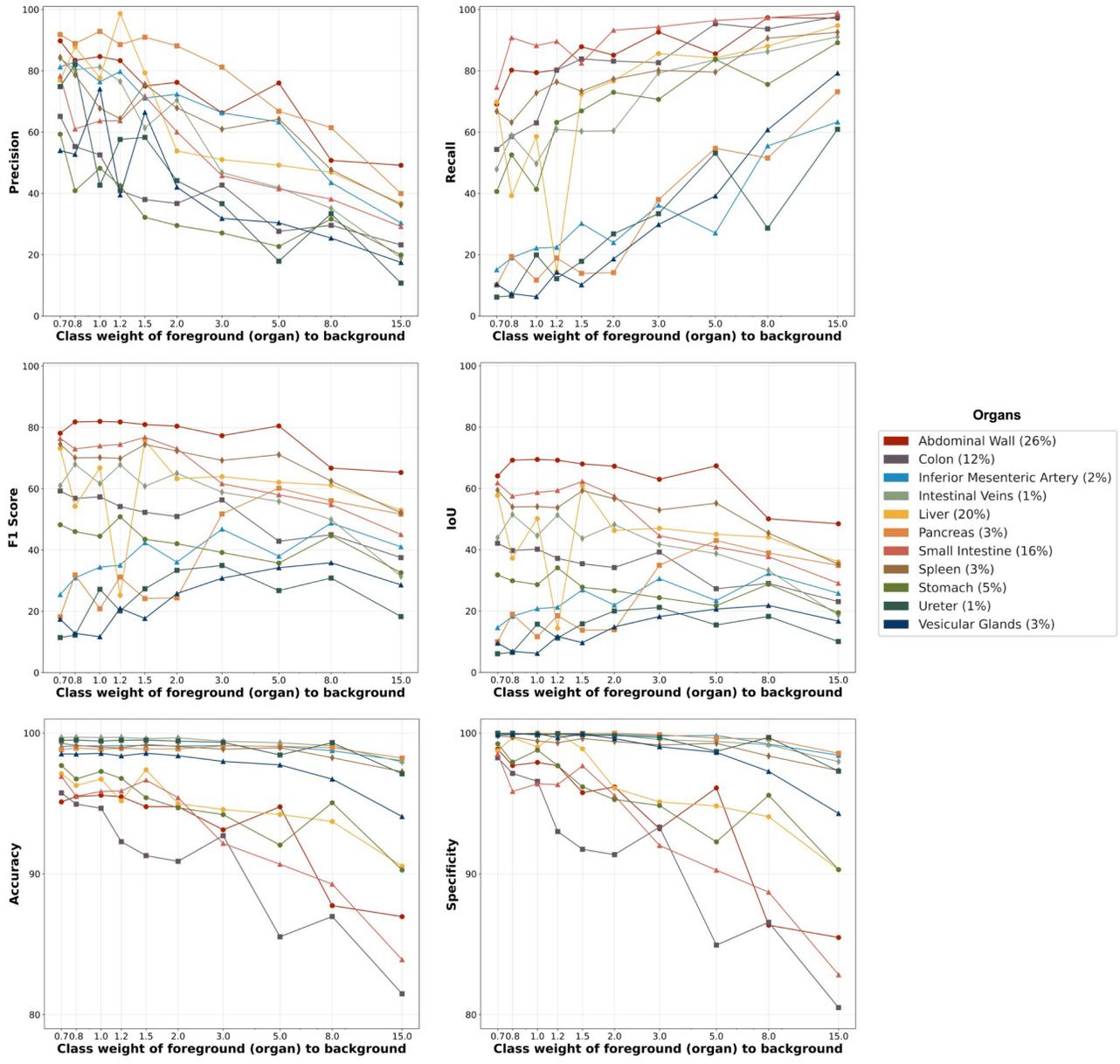

**Figure 3: Impact of variations in foreground-background class weights on segmentation performance metrics.** Models were trained and tested on supplemented training data including class-positive and class-negative examples. Abbreviation: Intersection-over-Union (IoU).

These class weight-related trends in counting metrics were similar when models were tested on organ-specific test data. Foreground-to-background class weight increases overall resulted in relative increase of distance-based segmentation performance metrics (HD and ASSD) (Supplementary Figure 2).

Optimal class weights for an organ differed between metrics. For example, pancreas segmentation performance was optimal in terms of precision when foreground-to-background class weight was 1.0, while the highest IoU and F1 score were observed at a foreground-to-background class weight of 5.0 (Figure 3). These observations highlight how class weight adjustments may help optimize segmentation model performance with regard to one or multiple metrics of interest, depending on the clinical use case.

## 5. Discussion

Accurate identification of risk and target structures is critically important to avoid complications and errors in surgery [48]. Consequently, organ segmentation is an important computer vision task in the field of surgical data science [2]. In this work, we illustrate the consequences of data bias related to (1) organ size and (2) representation of positive and negative examples in the training data on organ segmentation performance. Additionally, we demonstrate the impact of varying foreground-background class weights during the modeling process on eight segmentation performance metrics.

Our findings show that segmentation performance metrics are related to the size of the visible portion of the organ, with smaller organs exhibiting poorer performance compared to larger organs across most counting metrics. Furthermore, we quantify the effect of including negative examples in the training data: small organs particularly benefit from the inclusion of negative data in terms of improved precision, while larger organs in particular benefit from inclusion of negative training data in terms of accuracy and specificity. Last, we show how changes in class weights alter segmentation performance as determined by counting metrics and distance-based metrics. In conclusion, class weights can be adjusted to optimize specific metrics of interest, even at the expense of others.

Our work adds concrete evidence from the field of surgical data science to the ongoing discussion about selection and reporting of metrics in medical image analysis [37,39,41,49]. Specifically, our results highlight the distortive potential of poor selection of segmentation performance indicators and selective reporting of modeling parameters. As the surgical data science community anticipates an integration of computer vision models into intraoperative surgical decision support systems within the next decade [2,50], awareness of the impact of such errors is of utmost importance: In many surgical scenarios, blood vessels, for example, represent risk structures that need to be identified. False negatives (i.e., not identifying a visible blood vessel) may result in injury of the vessel and complications such as intraoperative bleeding. The target metric therefore needs to meet the demands of the specific clinical application scenario. In the example of risk structures such as a blood vessel that must not be violated, a metric penalizing false negatives, such as recall, would be more suitable than more general metrics such as IoU.

Our work has limitations. First, we analyzed only one dataset. While we expect the observed patterns to be transferable to other use cases beyond organ segmentation in laparoscopic imaging and similarly biased datasets, we do not prove this directly. Based on the limitation to the DSAD dataset, we only investigated the biases that this dataset represents. Biases such as a priori exclusion of patients with visibly diseased organs or different disease-related phenomena would need additional data and annotations that the DSAD dataset does not provide. Second, we selected a subset of eight commonly used metrics to evaluate segmentation performance. While these metrics are among those recommended for segmentation performance assessment, other metrics such as volumetric similarity or mean absolute surface distance may have provided additional insights [8]. Third, while the supplemented training dataset used in our experiments comes closer to reflecting clinical reality than the organ-specific data used in the baseline study [14], it represents a fixed distribution of images that display and do not display the class of interest. This does not fully model the true distribution of organ classes, which differs between surgery types. Last, our work focuses on binary segmentation problems and only considers one model architecture. In instance segmentation use cases, competition between classes may occur, resulting in more complex influences on nominal segmentation performance, which we did not explore in our study.

Despite these limitations, our work highlights the importance of critically evaluating and reporting data characteristics and algorithm parameters used in segmentation tasks. Ultimately, in the context of medical imaging, these modeling choices may be decisive for clinical applicability and patient safety.

## 6. Conclusion

In this work, we empirically investigate the impact of object size and class distribution in training and test data as well as class weights on popular segmentation performance metrics in the specific use case of anatomy segmentation in laparoscopic imaging. Our results indicate that changes in these parameters can introduce drastic changes in nominal performance of over 50%. Class weight alterations differentially impact segmentation performance metrics and thereby allow for regulation and optimization of individual model performance metrics. Overall, these findings provide insights into the impact of data biases on nominal model performance in surgical data science and support two adjustments to account for these biases: First, training on datasets that are similar to the clinical real-world scenarios in terms of class distribution, and second, class weight adjustments to optimize segmentation model performance with regard to metrics of particular relevance in the respective clinical setting.

# Strategies to Improve Real-World Applicability of Laparoscopic Anatomy Segmentation Models

Fiona R. Kolbinger
Purdue University
fkolbing@purdue.edu

Jiangpeng He
Purdue University
he416@purdue.edu

Jinge Ma
Purdue University
ma859@purdue.edu

Fengqing Zhu
Purdue University
zhu0@purdue.edu

# Supplementary Material

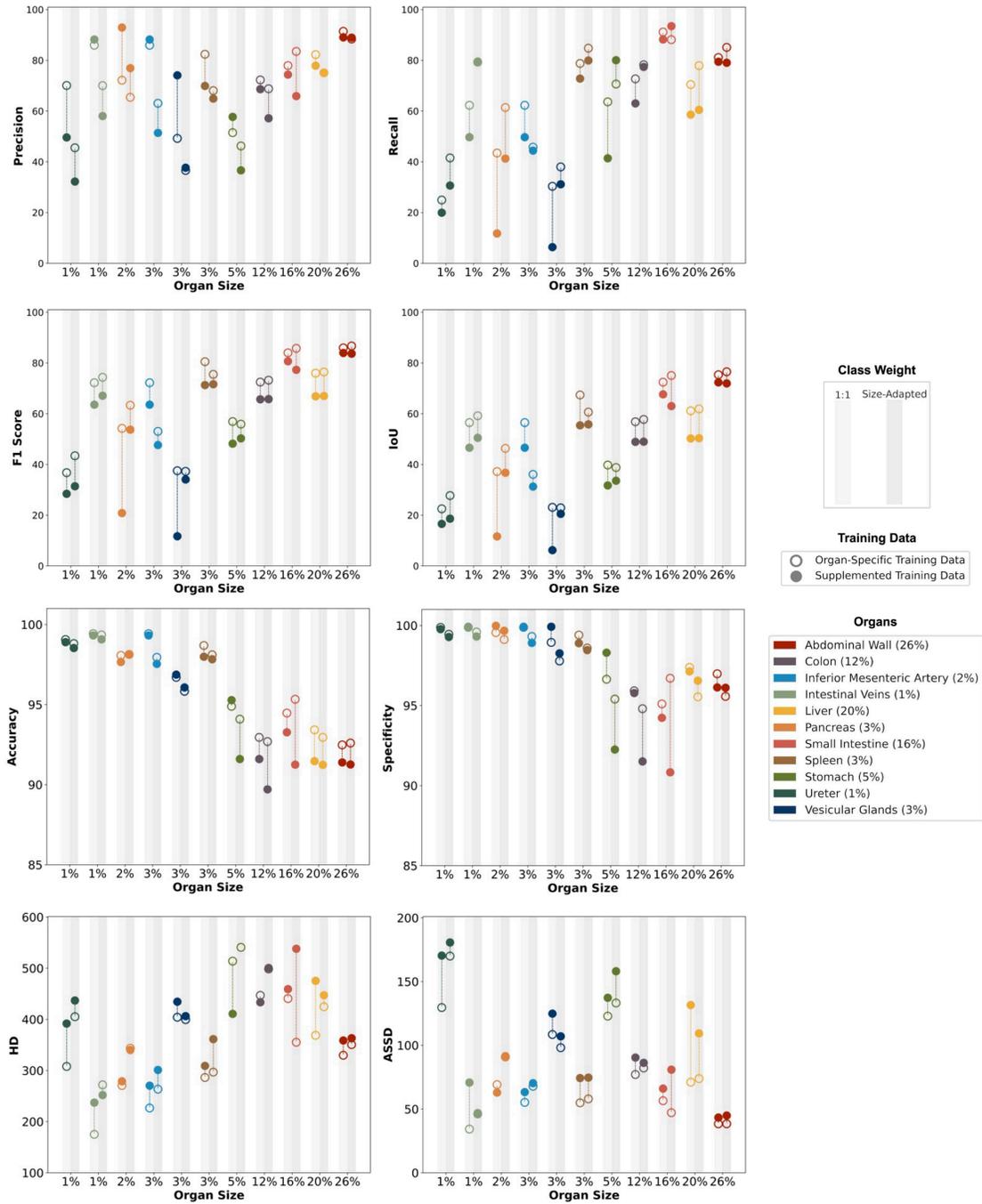

**Supplementary Figure 1: Quantitative evaluation of segmentation model performance in relation to training data composition and foreground-background class weight, tested on organ-specific test data.** Models were trained on either organ-specific training data including only class-positive images or supplemented training data including class-positive and class-negative examples. Foreground and background were either weighted identically (1:1) or adapted to organ size. Models were tested on organ-specific test data. Abbreviations: Average Symmetric Surface Distance (ASSD), Hausdorff Distance (HD), Intersection-over-Union (IoU).

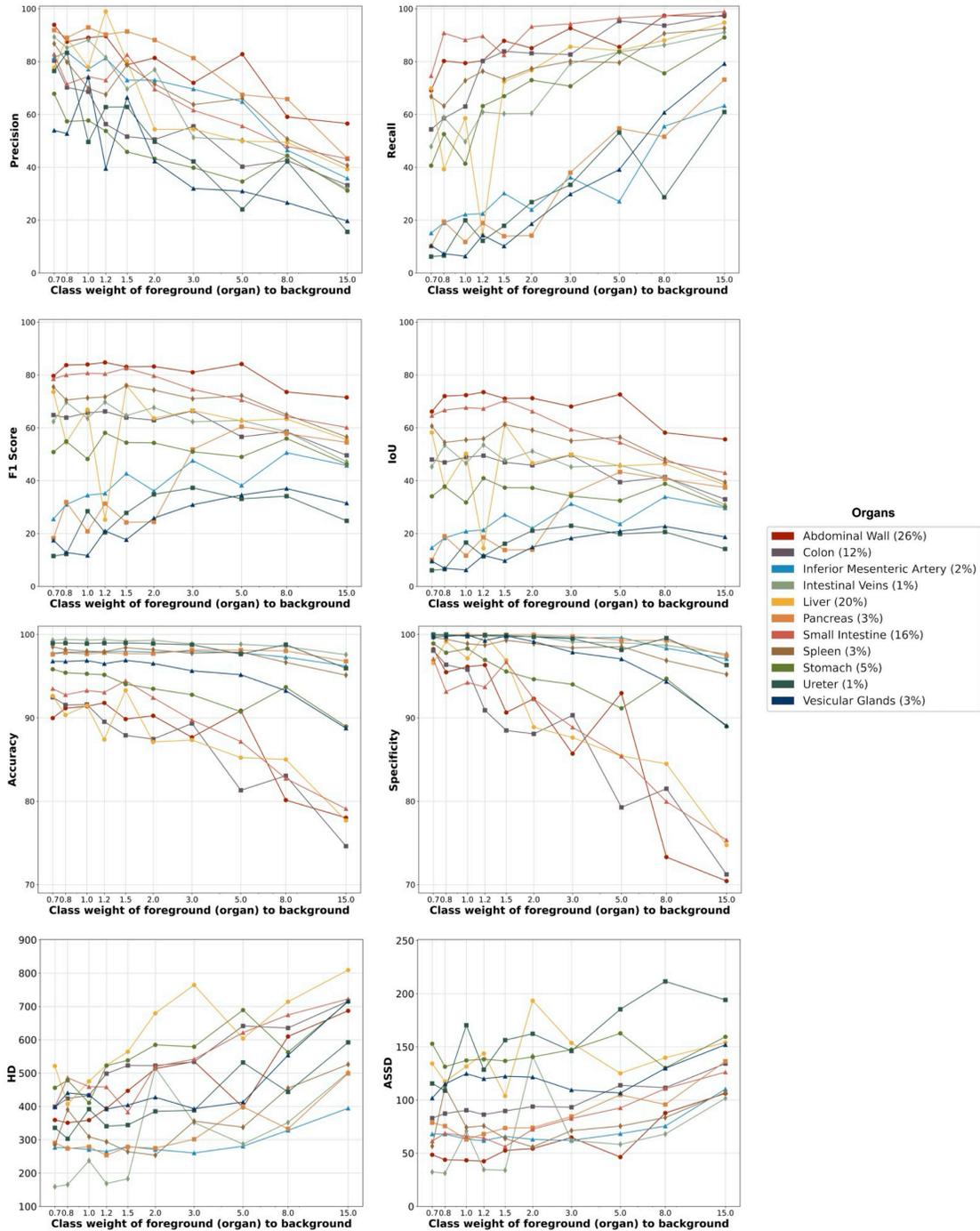

**Supplementary Figure 2: Impact of variations in foreground-background class weights on segmentation performance metrics, tested on organ-specific test data.** Models were trained on supplemented training data including class-positive and class-negative examples and tested on organ-specific test data. Abbreviations: Average Symmetric Surface Distance (ASSD), Hausdorff Distance (HD), Intersection-over-Union (IoU).

**Supplementary Table 1: Performance of binary organ segmentation models trained with class weights identical for foreground and background and tested on organ-specific test data.** Models were trained on organ-specific (O) training data including only class-positive images or supplemented (S) training data including class-positive and class-negative examples. All models were tested on organ-specific test data. Abbreviations: Average Symmetric Surface Distance (ASSD), Hausdorff Distance (HD), Intersection-over-Union (IoU), Organ-specific training dataset (O), Supplemented training dataset (S).

| Organ | Organ Size (% Foreground) | Accuracy | | Precision | | Recall | | IoU | | F1 Score | | Specificity | | HD | | ASSD | |
|---|---|---|---|---|---|---|---|---|---|---|---|---|---|---|---|---|---|
| | | O | S | O | S | O | S | O | S | O | S | O | S | O | S | O | S |
| Ureter | 1.2% | 99.06 | 98.90 | 70.07 | 49.60 | 24.92 | 19.94 | 22.52 | 16.58 | 36.76 | 28.45 | 99.88 | 99.77 | 307.90 | 391.73 | 129.60 | 170.35 |
| Intestinal Veins | 1.3% | 99.43 | 99.32 | 85.94 | 88.20 | 62.26 | 49.69 | 56.50 | 46.60 | 72.20 | 63.57 | 99.88 | 99.92 | 175.05 | 237.23 | 34.26 | 70.78 |
| Pancreas | 2.7% | 98.07 | 97.66 | 72.15 | 92.92 | 43.44 | 11.74 | 37.20 | 11.64 | 54.23 | 20.85 | 99.55 | 99.98 | 270.60 | 278.81 | 69.12 | 62.91 |
| Inferior Mesenteric Artery | 2.8% | 99.43 | 99.32 | 85.94 | 88.20 | 62.26 | 49.69 | 56.50 | 46.60 | 72.20 | 63.57 | 99.88 | 99.92 | 226.56 | 270.4 | 55.21 | 63.27 |
| Vesicular Glands | 2.9% | 96.70 | 96.87 | 49.22 | 74.08 | 30.33 | 6.36 | 23.10 | 6.22 | 37.53 | 11.71 | 98.94 | 99.92 | 404.08 | 434.42 | 108.47 | 124.82 |
| Spleen | 3.2% | 98.69 | 97.99 | 82.38 | 69.89 | 78.75 | 72.79 | 67.39 | 55.42 | 80.52 | 71.31 | 99.40 | 98.89 | 286.23 | 309.00 | 54.85 | 74.29 |
| Stomach | 5.0% | 94.89 | 95.28 | 51.48 | 57.70 | 63.62 | 41.38 | 39.77 | 31.75 | 56.91 | 48.20 | 96.64 | 98.30 | 513.78 | 410.70 | 122.87 | 137.24 |
| Colon | 11.8% | 92.95 | 91.60 | 72.26 | 68.61 | 72.68 | 63.01 | 56.83 | 48.91 | 72.47 | 65.69 | 95.92 | 95.78 | 446.84 | 433.33 | 77.08 | 90.38 |
| Small Intestine | 15.5% | 94.47 | 93.27 | 77.92 | 74.35 | 91.14 | 88.21 | 72.43 | 67.63 | 84.01 | 80.69 | 95.10 | 94.23 | 440.6 | 459.11 | 56.50 | 66.05 |
| Liver | 19.7% | 93.42 | 91.47 | 82.24 | 77.88 | 70.46 | 58.58 | 61.16 | 50.23 | 75.99 | 66.87 | 97.38 | 97.13 | 368.69 | 475.25 | 71.06 | 131.55 |
| Abdominal Wall | 26.2% | 92.48 | 91.39 | 91.41 | 89.04 | 81.10 | 79.44 | 75.35 | 72.36 | 85.95 | 83.97 | 96.98 | 96.13 | 329.73 | 358.67 | 38.37 | 43.32 |

**Supplementary Table 2: Performance of binary organ segmentation models trained with class weights adapted to organ size and tested on organ-specific test data.** Models were trained on organ-specific (O) training data including only class-positive images or supplemented (S) training data including class-positive and class-negative examples. All models were tested on organ-specific test data. Abbreviations: Average Symmetric Surface Distance (ASSD), Hausdorff Distance (HD), Intersection-over-Union (IoU), Organ-specific training dataset (O), Supplemented training dataset (S).

| Organ | Organ Size (% Foreground) | Accuracy | | Precision | | Recall | | IoU | | F1 Score | | Specificity | | HD | | ASSD | |
|---|---|---|---|---|---|---|---|---|---|---|---|---|---|---|---|---|---|
| | | O | S | O | S | O | S | O | S | O | S | O | S | O | S | O | S |
| Ureter | 1.2% | 98.81 | 98.53 | 45.51 | 32.22 | 41.52 | 30.65 | 27.73 | 18.63 | 43.43 | 31.41 | 99.45 | 99.28 | 405.10 | 436.90 | 169.96 | 180.67 |
| Intestinal Veins | 1.3% | 99.35 | 99.07 | 70.02 | 58.03 | 79.29 | 79.52 | 59.19 | 50.48 | 74.36 | 67.10 | 99.59 | 99.31 | 271.91 | 251.89 | 46.14 | 46.83 |
| Pancreas | 2.7% | 98.13 | 98.13 | 65.40 | 76.91 | 61.38 | 41.27 | 46.33 | 36.72 | 63.33 | 53.71 | 99.12 | 99.67 | 343.57 | 340.13 | 90.85 | 91.49 |
| Inferior Mesenteric Artery | 2.8% | 97.95 | 97.54 | 63.05 | 51.39 | 45.77 | 44.42 | 36.09 | 31.28 | 53.04 | 47.65 | 99.31 | 98.91 | 263.44 | 301.03 | 67.83 | 70.27 |
| Vesicular Glands | 2.9% | 95.83 | 96.07 | 36.58 | 37.66 | 37.98 | 31.13 | 22.90 | 20.54 | 37.27 | 34.09 | 97.78 | 98.26 | 399.46 | 406.33 | 98.11 | 107.02 |
| Spleen | 3.2% | 98.11 | 97.83 | 68.05 | 64.89 | 84.77 | 79.93 | 60.64 | 55.80 | 75.50 | 71.63 | 98.59 | 98.46 | 296.75 | 361.33 | 57.94 | 74.66 |
| Stomach | 5.0% | 94.09 | 91.60 | 46.24 | 36.65 | 70.65 | 80.04 | 38.79 | 33.58 | 55.90 | 50.28 | 95.40 | 92.25 | 540.69 | 656.05 | 133.18 | 158.10 |
| Colon | 11.8% | 92.69 | 89.71 | 68.79 | 57.16 | 78.25 | 77.39 | 57.75 | 48.98 | 73.22 | 65.76 | 94.80 | 91.51 | 498.50 | 500.49 | 82.19 | 86.24 |
| Small Intestine | 15.5% | 95.33 | 91.25 | 83.50 | 65.90 | 88.14 | 93.48 | 75.06 | 63.00 | 85.76 | 77.30 | 96.70 | 90.83 | 355.14 | 538.04 | 47.13 | 80.89 |
| Liver | 19.7% | 92.95 | 91.24 | 75.07 | 75.10 | 77.91 | 60.46 | 61.89 | 50.36 | 76.46 | 66.99 | 95.54 | 96.55 | 424.64 | 447.13 | 73.86 | 109.35 |
| Abdominal Wall | 26.2% | 92.60 | 91.26 | 88.38 | 88.93 | 85.09 | 79.03 | 76.53 | 71.95 | 86.70 | 83.69 | 95.57 | 96.11 | 350.55 | 363.21 | 38.39 | 44.88 |